\pdfoutput=1

\documentclass[11pt]{article}

\usepackage{emnlp2023}

\usepackage{times}
\usepackage{latexsym}

\usepackage[T1]{fontenc}

\usepackage[utf8]{inputenc}

\usepackage{microtype}

\usepackage{inconsolata}

\usepackage{mathtools}
\usepackage{times}
\usepackage{latexsym}

\usepackage{import}
\usepackage{hyperref}
\usepackage{graphicx}
\usepackage{adjustbox}
\usepackage{multirow}
\usepackage{verbatim}
\usepackage{algorithmic}
\usepackage{algorithm}
\usepackage{balance}
\usepackage{rotating}
\usepackage{xcolor}
\usepackage{enumitem}
\usepackage{footmisc}
\usepackage{caption}
\usepackage{subcaption}

\title{FaNS: a Facet-based Narrative Similarity Metric}

\author{Mousumi Akter, Shubhra Kanti Karmaker (``Santu'') \\
        Big Data Intelligence (BDI) Lab, Auburn University, Alabama, USA \\ 
        \{mza0170, sks0086\}@auburn.edu}

\begin{document}
\maketitle

\begin{abstract}
Similar Narrative Retrieval is a crucial task since narratives are essential for explaining and understanding events, and multiple related narratives often help to create a holistic view of the event of interest. To accurately identify semantically similar narratives, this paper proposes a novel narrative similarity metric called \textbf{Fa}cet-based \textbf{N}arrative \textbf{S}imilarity (FaNS), based on the classic 5W1H facets (Who, What, When, Where, Why, and How), which are extracted by leveraging the state-of-the-art Large Language Models (LLMs). Unlike existing similarity metrics that only focus on overall lexical/semantic match, FaNS provides a more granular matching along six different facets independently and then combines them. To evaluate FaNS, we created a comprehensive dataset by collecting narratives from AllSides, a third-party news portal. Experimental results demonstrate that the FaNS metric exhibits a higher correlation (\textbf{37\%$\uparrow$)} than traditional text similarity metrics that directly measure the lexical/semantic match between narratives, demonstrating its effectiveness in comparing the finer details between a pair of narratives.
\end{abstract}

\section{Introduction}
\begin{figure*}[!htb]
    \centering
    \includegraphics[width=\linewidth, ,trim={15 90 15 90},clip]{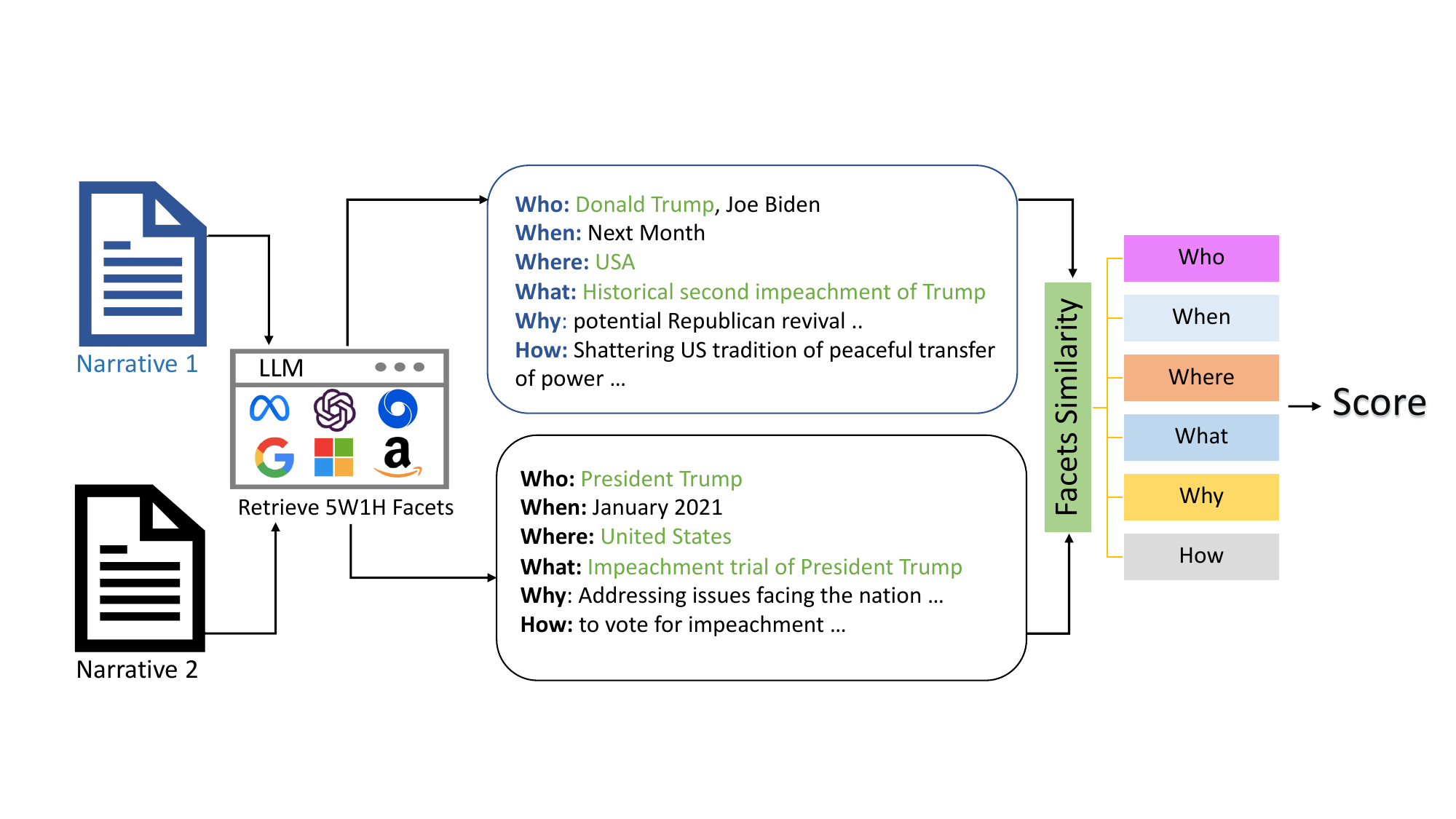}
    \caption{Computing \textbf{Fa}cet-based \textbf{N}arrative \textbf{S}imilarity (FaNS) with Large Language Models (LLMs)}
    \label{overview}
\end{figure*} 

% - overview 
Humans observe real-world events and report them in the form of narratives. Narratives are often complex, containing a variety of elements, such as characters, entities, and actions, that define the plot and provide continuity~\cite{mcintyre-lapata-2010-plot}. Yet, a single narrative is often incomplete and provides a biased/narrow view of the event of interest. To mitigate this issue, humans usually read multiple related narratives from different perspectives to perceive a holistic view of the event of interest. As humans read through these multiple related narratives, they can quickly draw connections between sub-events and important characters/entities, which helps them in having a more accurate and comprehensive understanding of the event of interest.

In this paper, we investigate the following research question: \textit{Given a narrative of an event of interest, how can we retrieve other similar/alternate narratives of that event?} One option is to use traditional textual similarity metrics like BERTScore~\cite{DBLP:conf/iclr/ZhangKWWA20}, ROUGE~\cite{lin-2004-rouge}, etc. However, one problem with these traditional metrics is that they only focus on the overall lexical/semantic match between two documents without giving special attention to the finer details of characters, entities, and actions~\cite{DBLP:conf/acl/AkterBS22,DBLP:journals/corr/abs-2308-02270}, which is very important for measuring the similarity between two narratives. In other words, the plain textual similarity metrics are often not capable of matching the granular elements between two narratives.

The question of how one should measure the similarity between a pair of narratives is indeed philosophical. Yet, one can safely argue that plain textual similarity metrics are definitely an oversimplified solution to the actual narrative similarity problem. Indeed, the characters, their relations, and their actions are often more important parts of a narrative than other circumstantial details, and a \textit{``good''} narrative similarity metric should give higher importance to matching characters/actions rather than matching circumstantial details. In fact, \citet{DBLP:journals/pieee/XieSC08} argued that an event can be comprehensively represented by the following six basic facets: Who, When, Where, What, Why, and How, often referred to as 5W1H notation. For example, one can characterize an event of interest by asking the following questions related to the event: \textit{Who are the major characters in the narrative/event? When did this event occur? Where did this event occur? What happened? What is the discussion's main point? How and why did this occur?} Therefore, a reasonable way to compute the similarity between a pair of narratives is to ask the same set of 5W1H questions to both narratives and compare the answers for each question. This is essentially the approach we adopted for our FaNS metric.

%A great deal of work has been done centered on just one facet of the narrative/event, such as character (Who)~\cite{DBLP:conf/emnlp/BrahmanHTZSC21} or event-centric~\cite{DBLP:conf/acl/ChambersJ08,wang-etal-2022-uncovering} narrative understanding. It is a very tedious and time-consuming task to ask humans to structure narratives according to these 5W1H facets; perhaps that's why there hasn't been any work that combines these essential questions for comprehending a narrative rather than focusing single aspect only so far~\cite{huang-etal-2019-cosmos, DBLP:conf/emnlp/LalTALCMB22}. However, 

A major challenge for the FaNS metric is accurately extracting the 5W1H facets from two input narratives and performing a facet-wise similarity computation to derive an overall similarity score. Thanks to the recent development of Large Language Models (LLMs)~\cite{DBLP:conf/nips/KojimaGRMI22,DBLP:conf/nips/Ouyang0JAWMZASR22,DBLP:journals/corr/abs-2304-13712}, we can now leverage the power of these LLMs to precisely extract the 5W1H facets from narratives, which was quite challenging even a few years ago. More specifically, to implement FaNS, we extracted the 5W1H facets using two popular LLMs, i.e., ChatGPT~\cite{DBLP:journals/corr/abs-2305-13300} and Bard\footnote{\href{https://bard.google.com/}{https://bard.google.com/}}. Next, we computed individual facet-based similarity scores to determine the similarity along each facet between the two input narratives. Finally, using a variety of aggregation techniques (such as a linear combination of entity-specific and descriptive facets or a custom-weighted average of individual facets), we integrated all of these facet-based similarities to derive a final score. Figure~\ref{overview} shows this process.

For evaluation, we used news articles from the AllSides\footnote{\href{https://www.allsides.com/}{https://www.allsides.com/}} data set introduced by~\citet{DBLP:conf/coling/BansalAS22}. The Allsides website organizes alternative narratives on a particular event/topic reported by different politically biased news media and also provides a so-called ``neutral'' description of their own which they call as \textit{theme}. Additionally, Allsides provides related topic tags for each event; however, it does not provide human annotations for narrative similarity, which is important for the meta-evaluation of any metric. To address this issue, we came up with an intuitive yet innovative way of creating multiple ground truth labels for characterizing the pairwise narrative similarity based on the event and topics provided by Allsides. 

%assigned multiple labels to various news pairs from Allsides based on similarity with events/topics/topic groupings which we used as ground truth.

Note that, in this work, we first extract facets using LLMs and then compute their similarities and combine them rather than directly asking LLMs to provide a similarity score given two narratives as inputs during the prompt. We made this choice because receiving a score directly from the LLM immediately transforms the metric into a ``black box'' making it challenging to figure out what the number really means. Instead, we first structure both narratives along the 5W1H facets using LLMs and then derive an intuitive metric that provides greater control and better explainability. Finally, We conducted extensive experiments with the FaNS metric using AllSides data. Our experiments show that traditional metrics like ROUGE and BERTScore, which directly assess the overall lexical/semantic overlap, have a lower correlation (by 37\% less) against the ground truth labels than the \textbf{Fa}cet-based \textbf{N}arrative \textbf{S}imilarity (FaNS)  metric. In summary, this paper presents several notable contributions:

\begin{itemize}[leftmargin=*,itemsep=0.2ex,partopsep=0.2ex,parsep=0.2ex]
    
    \item We propose a facet-based narrative similarity metric, i.e., FaNS, which compares a pair of narratives along the classic 5W1H facets (Who, What, When, Where, Why, and How) to quantify the overall similarity between the input narratives.

    \item We curate and contribute a comprehensive dataset exclusively for the meta-evaluation of narrative similarity metrics (existing and future).
    
    \item We demonstrate that our facet-based narrative similarity (FaNS) metric achieves a higher correlation (\textbf{37\%$\uparrow$}) against the ground truth labels than traditional textual similarity metrics. 
\end{itemize}

\section{Related Work}
An accurate evaluation of narrative/document similarity has always been challenging~\cite{novikova-etal-2017-need,wang-etal-2018-metrics}. The common practice for computing the similarity of a pair of narratives/documents has been to compare the overall lexical (n-gram-based)/semantic (embeddings-based) overlap between those narratives/documents. For example, ROUGE~\cite{lin-2004-rouge} considers direct lexical overlap, while metrics like S+WMS~\cite{DBLP:conf/acl/ClarkCS19}, MoverScore~\cite{DBLP:conf/emnlp/ZhaoPLGME19}, and BERTScore~\cite{DBLP:conf/iclr/ZhangKWWA20}, are based on semantic similarity between two documents.

Recently, researchers have spent a lot of effort evaluating different aspects of text generation and summarization techniques that rely on measuring textual similarity in some capacity. For example, \citet{DBLP:journals/tacl/FabbriKMXSR21, DBLP:conf/emnlp/Zhong0YMJLZJH22} discussed how to perform meta-evaluation of summarization metrics along four explainable dimensions: \textit{coherence, consistency, fluency, and relevance}. Different question-answering frameworks have been proposed to evaluate the factual consistency of summaries~\cite{DBLP:conf/acl/WangCL20,DBLP:conf/acl/DurmusHD20,DBLP:conf/emnlp/ScialomDLPSWG21,DBLP:conf/naacl/FabbriWLX22}.

As an interesting development, recent research has witnessed the emergence of Large Language Models (LLMs) like ChatGPT~\cite{DBLP:journals/corr/abs-2305-13300}, as a versatile tool for evaluating various NLP tasks as well. For instance, \citet{DBLP:journals/corr/abs-2304-02554, DBLP:journals/corr/abs-2302-04166, DBLP:journals/corr/abs-2305-14540, DBLP:journals/corr/abs-2303-04048} investigated the strengths and limitations of ChatGPT as an evaluator of textual similarity and summarization quality. ChatGPT has also been used as an evaluator of factual consistency of abstractive text summarization techniques~\cite{DBLP:journals/corr/abs-2303-15621}. Moreover, the GPT-4 model, as an evaluator of text generation and similarity tasks, shows better alignment with human judgement~\cite{DBLP:journals/corr/abs-2303-16634}.

Still, None of these works focus on a facet-based narrative similarity metric that organizes input narratives along the 5W1H facets and combines individual facet-based similarity scores. The majority of works on direct evaluation using LLMs are ``black boxes'', and users are generally unaware of why they obtain such a score and what to improve. Most prior research has also concentrated mostly on summary evaluation or text generation evaluation by comparing the machine-generated text against a human-written reference text in a holistic fashion. In contrast, the focus of our work is on narrative similarity (irrespective of whether machine-generated or human written), where individual facets play an important role in determining the similarity scores, and hence, traditional metrics can not be applied directly.

\section{5W1H Facets of Narratives}
A narrative is essentially a portrayal of human perception of a real-world event via natural language description. Theoretically, each event can be summarized by answering the following six questions related to the event of interest: \textit{Who? When? Where? What? Why? How?}, which are defined as the 5W1H facets of an event. \citet{DBLP:journals/pieee/XieSC08} argued that these 5W1H facets provide sufficient and necessary information to summarize an event.

\vspace{-2mm}
\begin{itemize}[leftmargin=*,itemsep=0ex,partopsep=0ex,parsep=0ex]
 
\item{\textbf{Who}}: Main subjects of the event being described in a narrative. While the ``Who'' facet mainly refers to actual individuals or characters involved in an event, it may occasionally refer to an object.

\item{\textbf{When}}: Time when the event actually took place.

\item{\textbf{Where}}: Different places or specific geographic locations where the event took place.

\item{\textbf{What}}: A highly concise summary of the event, including actions, activities, and consequences.

\item{\textbf{Why}}: A brief explanation of the context of the event, its causes, and related background.

\item{\textbf{How}}: Some specific details on the actual event of interest, e.g., numbers, statistics, etc.
\vspace{-2mm}
\end{itemize}

The 5W1H facets help provide a clear picture of an event and can uncover hidden insights that are often overlooked when examining narratives as a whole. Table~\ref{tab:ex_facet} provides an example of using 5W1H facets to summarize an event narrative with the help of LLMs, e.g., ChatGPT. Both narratives, which are provided by a left-wing news media and a right-wing news media, focus on a Middle East issue and President Trump's involvement in it. It is evident from this structured summary representation that the left and right-wing news media addressed the \textit{Why} component in their narratives differently. As such, comparing the 5W1H facets between two narratives provides a robust way of measuring the similarity between them while facilitating interpretation and explanation as well. In this paper, we primarily adopted zero-shot approaches~\cite{DBLP:conf/emnlp/SarkarFS23,DBLP:conf/ijcnlp/SarkarFS22} while prompting LLMs for simplicity.

\begin{table*}[!htb]
\footnotesize
\centering
\resizebox{\textwidth}{!}{
\begin{tabular}{ll}
\hline
\multicolumn{2}{l}{\begin{minipage}[t]{\textwidth}%
\centering
\textbf{Topic Group:} National Security and Defense \hspace{1 cm} \textbf{Topic:} Middle East
\end{minipage}} \\ \hline
\multicolumn{2}{l}{\begin{minipage}[t]{\textwidth}%
\centering
\textbf{From Left News}
\end{minipage}} \\ \hline
\multicolumn{2}{l}{\begin{minipage}[t]{\textwidth}%
\textbf{Title}: The \textcolor{blue}{Israel}-UAE agreement is an insult to the peace Palestinians and Arabs want and need\\
\textbf{Narrative:} On \textcolor{green}{Thursday}, \textcolor{red}{President Trump} announced that \textcolor{red}{U.S. diplomats} had brokered a major breakthrough. The agreement basically declares that the corrupt government of \textcolor{red}{Israeli Prime Minister Benjamin Netanyahu} will postpone its plans for the unilateral annexation of occupied Palestinian land in return for normalization with a small but rich \textcolor{blue}{Arab} Gulf state. The problem with the big hoopla around this is that it is neither a genuine breakthrough nor will it bring peace between Israelis and \textcolor{blue}{Arabs} anytime soon.
... ... ...
Had Israel gone through with its unilateral annexation plans, it would have jeopardized its peace agreements with Jordan and Egypt and could have brought sanctions from traditional European supporters such as Germany, Britain and France. The tragedy in all of this is that \textcolor{red}{Palestinians} will be totally erased from all the talk of “breakthroughs” and “peace.” But we know that genuine peace will only come by seriously committing to end the Israeli occupation and creating a democratic and peaceful Palestinian state that can exist side-by-side a safe and secure Israel. Until then, all will be just part of an insulting diplomatic circus.
\vspace{1 mm}%
\end{minipage}} \\ \hline 

\multicolumn{2}{l}{\begin{minipage}[t]{\textwidth}%
\textbf{Who:} \textcolor{red}{President Trump}, \textcolor{red}{U.S. diplomats}, \textcolor{red}{Israeli Prime Minister Benjamin Netanyahu}, \textcolor{red}{Palestinians} \\
\textbf{When:} \textcolor{green}{Thursday}, 1967, August 13 \\
\textbf{Where:} \textcolor{blue}{Arab lands}, \textcolor{blue}{Israel} \\
\textbf{What:} The US-brokered agreement postponed Israeli annexation of Palestinian land in exchange for normalization with a Gulf state but fails to provide a road map for ending the decades-long Israeli occupation of Palestinian lands.\\
\textbf{Why:} Netanyahu and Trump hoped to improve their political positions, and Crown Prince Sheikh Mohammed bin Zayed needed a diplomatic victory. Genuine peace will only come by committing to end the Israeli occupation and creating a democratic Palestinian state.\\
\textbf{ How: }The US brokered an agreement between the Gulf state and Israel to postpone annexation but failed to address the decades-long Israeli occupation of Palestinian lands.
\vspace{1 mm}%
\end{minipage}} \\ \hline

\multicolumn{2}{l}{\begin{minipage}[t]{\textwidth}%
\centering
\textbf{From Right News}
\end{minipage}} \\ \hline
\multicolumn{2}{l}{\begin{minipage}[t]{\textwidth}%
\textbf{Title}: Trump’s Mideast Breakthrough\\
\textbf{Narrative:} \textcolor{red}{President Trump}’s Mideast strategy has been to strongly back \textcolor{red}{Israel}, support the Gulf monarchies, and press back hard against Iranian imperialism. His liberal critics insisted this would lead to catastrophe that never came, and on \textcolor{green}{Thursday} it delivered a diplomatic achievement: The \textcolor{red}{United Arab Emirates} and Israel agreed to normalize relations, making the \textcolor{red}{UAE} the first Arab League country to recognize the Jewish state in 20 years.
... ... ...
With the \textcolor{red}{UAE} deal, \textcolor{red}{Mr. Netanyahu} can avoid annexation while protecting against criticism from his right. The \textcolor{red}{UAE} can say it blocked annexation and protected the Palestinian cause. But the fact that annexation was a bargaining chip at all shows how the balance of power in the Israel-Palestine conflict has shifted in Israel’s favor. Arab states would previously have demanded far greater concessions in exchange for recognition. But the Iran threat, plus the Palestinians’ long-running rejectionism, has made that issue less important to Arab states.
\vspace{1 mm}%
\end{minipage}} \\ \hline
\multicolumn{2}{l}{\begin{minipage}[t]{\textwidth}%
\textbf{Who:} \textcolor{red}{President Trump}, \textcolor{red}{Israel}, \textcolor{red}{United Arab Emirates}, \textcolor{red}{Mr. Netanyahu} \\
\textbf{When: }\textcolor{green}{Thursday} \\
\textbf{Where:} Middle East\\
\textbf{What:} United Arab Emirates and Israel have agreed to normalize relations \\
\textbf{Why: }To strengthen regional checks on Iranian power and protect against criticism\\
\textbf{How: }Mr. Trump's pivot from Iran reassured Israel and Gulf states and put the U.S. in a position to broker agreements
\vspace{1 mm}%
\end{minipage}} \\ \hline

\end{tabular}
}
\caption{An example of narrative organization utilizing 5W1H facets for two news articles from Allsides}
\label{tab:ex_facet}
\end{table*}

\section{Computing FaNS Metric}

\subsection{Extracting 5W1H Facets Using LLMs}
\begin{figure*}[!htb]
    \centering
    \includegraphics[width=0.95\linewidth, ,trim={0 0 0 0},clip]{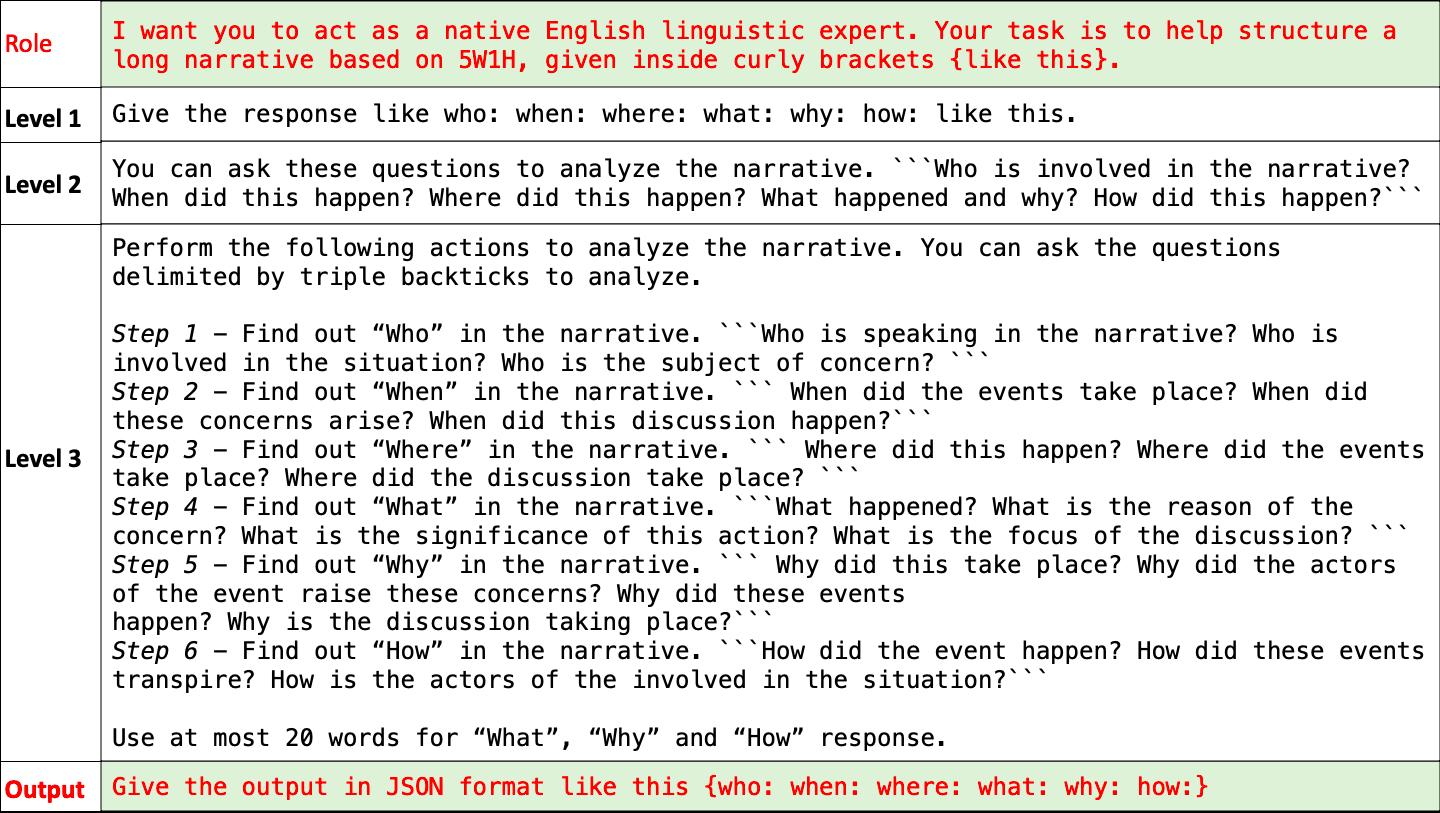}
    \caption{Various levels of prompts used to retrieve 5W1H facets}
    \label{fig:prompt}
\end{figure*}

Recently, LLMs have made tremendous advancements in their capacity to perform a variety of NLP tasks, along with performing as an evaluator of those tasks. For example, ChatGPT has been utilized to assess the quality of model-generated summaries directly. Also, such evaluations have been compared against numerous qualitative assessments and demonstrated a higher correlation with human judgments~\cite{DBLP:journals/corr/abs-2301-13848}. In contrast, for our FaNS metric, we first compare the structured representations of two narratives along the 5W1H dimensions and then combine them into a final score. To achieve this, we prompted LLMs to extract 5W1H facets from narratives rather than asking them to provide similarity scores directly.

With regard to prompting LLMs, multiple research works have repeatedly emphasized that careful prompt design/engineering is very important for effectively using LLMs~\cite{DBLP:journals/corr/abs-2211-01910}. Very recently, the TELeR~\cite{DBLP:journals/corr/abs-2305-11430} taxonomy was proposed, which demonstrated how to design and benchmark LLM prompts for complex tasks. To extract the 5W1H facets from input narrative pairs, we followed the TELeR taxonomy to design three levels of prompts, i.e., Levels 1, 2, and 3. More specifically, we first defined the role of LLM as a linguist who is a native English speaker. Then, given a narrative, we instructed the LLM  to organize the input narratives based on 5W1H facets. Here, Level 1 means a very high-level prompt with just a single sentence, while in Level 2, we  provided more detailed instructions in a paragraph style, including the distinct sub-tasks that need to be performed. We gave more thorough instructions and defined six distinct steps as sub-tasks in Level 3, with each step describing how to extract each facet. Figure~\ref{fig:prompt} illustrates the different levels of prompts we designed as part of this work.

%We hypothesized that because Level 3 prompts are more detailed, the LLM will understand these prompts better~\cite{DBLP:conf/nips/Wei0SBIXCLZ22}, and therefore, LLM's response to Level 3 prompts will be better than that of levels 1 and 2. 

%The 5W1H facets, each indicating a different aspect of the narrative, allow for a more accurate comparison of narratives. Section~\ref{facet_sim} provides a thorough methodology for assessing narrative similarity and outlines the intuition behind the metric design for each facet's similarity.

\subsection{Matching Individual Facets}

Below, we describe how the similarity for each individual facet (extracted by LLMs) was assessed.

{\textbf{Comparing \textit{Who}:}} This facet examines whether the input narratives contain common characters/ actors or not. We used both fuzzy string matching and entity-embedding-based matching to assess the similarity among the characters/ actors. For the fuzzy string matching, if there is at least 80\% overlap, we considered this as a match. If not, we then used entity embeddings to compute the cosine similarity  between character entities and, based on a user-defined threshold, determined whether a match had indeed occurred. After finding out the common characters/ actors mentioned in both narratives, we computed precision and recall scores and the corresponding $F_1$ score, which was considered as the similarity for the ``Who'' facet.

%computing the overlap match between narratives one and two, we computed the precision with respect to narrative one and the recall with respect to narrative two. This precision and recall score are combined to get an F1 score.

{\textbf{Comparing \textit{When}:}} This facet looks at how the input narratives are timed, including time periods, timestamps, and the temporal connections between narratives. In order to compare the when facets, we used the Python Daytime Parser\footnote{\href{https://dateutil.readthedocs.io/en/stable/parser.html}{https://dateutil.readthedocs.io/en/stable/parser.html}}, which allows us to extract various ``When'' patterns, such as date patterns (MM/DD/YYYY or DD/MM/YYYY), day patterns (Monday through Sunday), month patterns (January through December), time patterns (AM/PM), and daytime patterns (Morning/Afternoon/Evening/Night). Based on the common timeframes mentioned in both narratives, we again computed the $F_1$ score as a similarity metric for the ``When'' facet.

%Then, given narratives one and two, we compute an overlap match. Following the computation of precision and recall with regard to narratives one and two, the F1 score is calculated.

%The geographic details, descriptions, and general spatial aspects of the narrative pairs can be compared to determine how similar they are.
%While comparing the \textit{Where} facet between two narratives, we observe sometimes LLMs provided whole sentences containing the location instead of mentioning the locations only. ven after performing lexical and semantic matching, it was crucial to compare the \textit{where} as one narrative may discuss any state of the USA (For example, Florida only) while another may refer to the United States as a whole in the narrative. Five is the maximum allowed from manual assessment.

{\textbf{Comparing \textit{Where}:}} This facet examines the spatial context of input narratives, including place markers and geographic locations. To compare locations, we first extracted the geopolitical, locational, facility, and organizational entities, including representative phrases of that entity using spaCy\footnote{\href{https://spacy.io/api/doc}{https://spacy.io/api/doc}} which we defined as the locations set and then we match the location sets between two narratives. To detect lexical matches, we used fuzzy string matching. If the fuzzy matching score is less than 80\%, then we generated entity embedding using spaCy NER\footnote{\href{https://spacy.io/usage/embeddings-transformers}{https://spacy.io/usage/embeddings-transformers}} to do semantic matching by computing cosine similarity with a threshold of 0.75. To further enhance our matching technique, we used ConceptNet~\cite{DBLP:conf/aaai/SpeerCH17} to extract the top five concepts that each location was related to and compare whether there is an overlap between the concepts of the mentioned locations. We next calculated the $F_1$ score based on the matching heuristics discussed so far.

{\textbf{Comparing \textit{What/Why/How}:}} These facets delve into brief descriptions of the core event, its underlying causes, and key specifics described in a narrative. We used the SEM-$F_1$~\cite{DBLP:conf/emnlp/BansalAS22} metric to quantify the similarity between these components since it has been reported to have higher inter-rater reliability and correlation with human judgment for textual overlaps and can be directly applied to our scenario. In order to compute the SEM-$F_1$ scores, we used the STSb-roberta-large~\cite{DBLP:conf/emnlp/ReimersG19} model to generate sentence embeddings for each facet and computed cosine similarity scores between them.

\subsection{Final Metric for Narrative Similarity}\label{facet_sim}

%We can add up all of the F1 scores we receive for the individual facets to reach a final score. 

We opted for two methods in order to determine the final similarity score. For the first method, we divided the facets into two categories: entity-specific facets (who, when, and where) and descriptive facets (what, why, and how). Next, we computed a weighted average of the entity-specific facet similarity scores (with weight $\alpha$) and the descriptive facet similarity scores (with weight $1-\alpha$) to compute the final FaNS metric. Additionally, we adopted a separate weighted linear combination technique as our second approach, with weights derived by statistical correlation (details in Section~\ref{entity_descrip} and~\ref{facet_weight}). 

\section{Experiments and Results}
\subsection{Dataset}\label{dataset}
\setlength{\textfloatsep}{0pt}
\begin{figure*}[!htb]
    \centering
    \begin{subfigure}[b]{0.7\textwidth}
        \centering
        \includegraphics[width=\linewidth, trim={0 0 0 0}, clip]{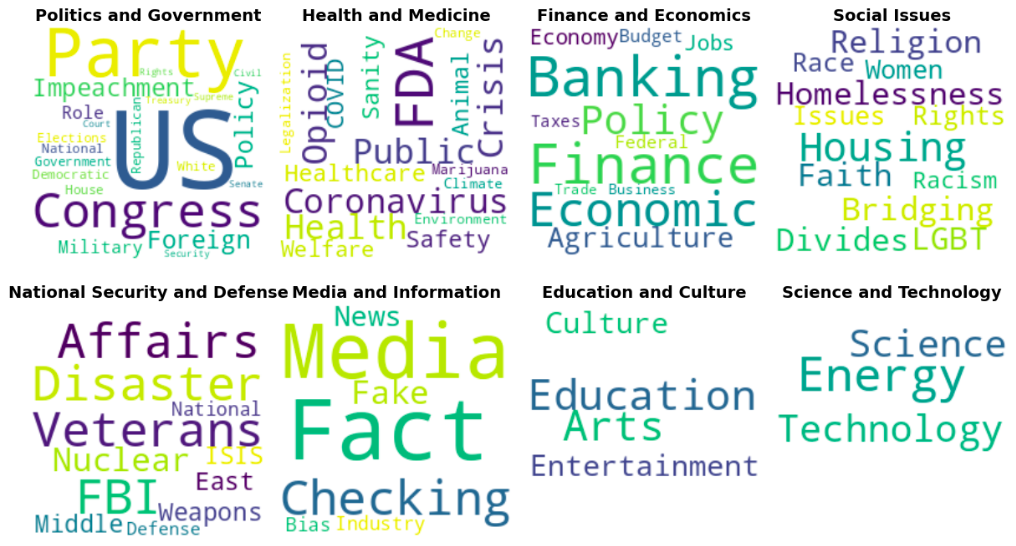}
    \end{subfigure}
    \caption{A visual illustration of numerous topic groups found in the Allsides data}
    \label{fig:topic_group}
\end{figure*}

We didn't find an existing dataset that we could readily use for the meta-evaluation of our narrative similarity metric, which provides ground truth similarity labels for a large number of narrative pairs. Existing information retrieval (IR) corpora with relevance labels for query-document pairs can not be used in our case because IR queries are often very short and do not contain the 5W1H facets most of the time. To address this challenge, we collected news articles from~\href{https://www.allsides.com/unbiased-balanced-news}{Allsides.com} introduced by~\citet{DBLP:conf/coling/BansalAS22}. For an event of interest, Allsides groups news articles from various sources based on their political leanings (left, right, center) and provides a ``neutral'' description of the event called \textit{theme}. Additionally, Allsides employs human annotators to cluster different events into specific topics. However, Allsides does not provide ground truth scores for narrative similarity, which is crucial for meta-evaluation. To solve this issue, we opted for an intuitive yet creative approach.

First, we created different topic groups, such as \textit{Finance and Economics}, which encompassed topics like banking and finance, economic policy, agriculture, and others. Figure~\ref{fig:topic_group} represents a visual representation of various topic groups created from Allsides narrative data. Next, we defined narrative similarity as a multi-label ordinal variable and heuristically assigned numeric scores to narrative pairs that we later used as ground truth in our meta-evaluation. Our heuristic was that narratives belonging to the same event should exhibit the highest similarity and were assigned a score of 3. Since events are grouped based on similarity by human annotators, narratives within the same topic but describing different events were assigned a similarity score of 2. Narratives from different topics within the same topic group were assigned a score of one, while narratives from different topic groups were assigned a score of zero. In summary, our heuristic is based on the following intuition: same event similarity $>$ same topic but different event similarity $>$ same topic group but different topic similarity $>$ others, which is very reasonable.

We randomly selected 205 events (including narratives from left, right, center, and also the theme) of around 100 topics from the Allsides data set. For each of the 205 events, we passed narratives from left, right, center, and theme, in total ($205 \times 4 = 820$) narratives from different topic groups and topics to LLM using Level 3, Level 2, and Level 1 prompts to retrieve the 5W1H facets. We then took all pairs of narratives and assigned them ground truth similarity labels based on our heuristic approach. Table~\ref{tab:data_stat} represents the counts of narrative pairs we used to calculate the correlation between the ground truth and the FaNS metric using ChatGPT as the LLM\footnote{Experiments with Google Bard is included in Appendix}. It is important to note that the samples for different levels may not always be identical due to various limitations, such as failures of LLM to retrieve facets (detail in Section~\ref{llm_fail}), or the LLM being overloaded with other requests, factors over which we had no control.

\begin{table}[!htb]\small
\centering
\resizebox{0.45\textwidth}{!}{
\begin{tabular}{cc|cc|cc}
\hline
\multicolumn{2}{c|}{\textbf{Level 1}} & \multicolumn{2}{c|}{\textbf{Level 2}} & \multicolumn{2}{c}{\textbf{Level 3}} \\ \hline
\multicolumn{1}{l|}{\textbf{Score}} & \multicolumn{1}{l|}{\textbf{Pairs}} & \multicolumn{1}{l|}{\textbf{Score}} & \multicolumn{1}{l|}{\textbf{Pairs}} & \multicolumn{1}{l|}{\textbf{Score}} & \multicolumn{1}{l}{\textbf{Pairs}} \\ \hline
\multicolumn{1}{c|}{3} & 669 & \multicolumn{1}{c|}{3} & 639 & \multicolumn{1}{c|}{3} & 669 \\ 
\multicolumn{1}{c|}{2} & 657 & \multicolumn{1}{c|}{2} & 629 & \multicolumn{1}{c|}{2} & 657 \\ 
\multicolumn{1}{c|}{1} & 700 & \multicolumn{1}{c|}{1} & 700 & \multicolumn{1}{c|}{1} & 700 \\ 
\multicolumn{1}{c|}{0} & 700 & \multicolumn{1}{c|}{0} & 700 & \multicolumn{1}{c|}{0} & 700 \\ \hline
\end{tabular}
}
\caption{Statistics of the narrative pairs used to calculate the correlation between the ground truth labels and the FaNS metric, where facets were extracted by ChatGPT.}
\label{tab:data_stat}
\vspace{-2mm}
\end{table}

\subsection{Failures of LLMs for Facet Retrieval}\label{llm_fail}
\begin{table}[!htb]
\centering
\resizebox{0.48\textwidth}{!}{
\begin{tabular}{l|cc|cc|cc}
\hline
\multirow{2}{*}{} & \multicolumn{2}{c|}{\textbf{Level 1}} & \multicolumn{2}{c|}{\textbf{Level 2}} & \multicolumn{2}{c}{\textbf{Level 3}} \\ \cline{2-7} 
 & \multicolumn{1}{c|}{\textbf{ChatGPT}} & \textbf{Bard} & \multicolumn{1}{c|}{\textbf{ChatGPT}} & \textbf{Bard} & \multicolumn{1}{c|}{\textbf{ChatGPT}} & \textbf{Bard} \\ \hline
\textbf{Left} & \multicolumn{1}{c|}{205} & 154 & \multicolumn{1}{c|}{188} & 154 & \multicolumn{1}{c|}{205} & 138 \\ 
\textbf{Right} & \multicolumn{1}{c|}{205} & 171 & \multicolumn{1}{c|}{203} & 177 & \multicolumn{1}{c|}{205} & 160 \\ 
\textbf{Center} & \multicolumn{1}{c|}{181} & 152 & \multicolumn{1}{c|}{181} & 161 & \multicolumn{1}{c|}{181} & 161 \\ 
\textbf{Theme} & \multicolumn{1}{c|}{205} & 194 & \multicolumn{1}{c|}{205} & 201 & \multicolumn{1}{c|}{205} & 200 \\ \hline
\textbf{Success} & \multicolumn{1}{c|}{97\%} & 81.8\% & \multicolumn{1}{c|}{94.8\%} & 84.5\% & \multicolumn{1}{c|}{97.1\%} & 80.4\% \\ \hline
\end{tabular}
}
\caption{Statistics of the samples we finally got after facet retrieval for ChatGPT and Bard for 205 events, including narratives from Left, Right, Center, and Theme.}
\label{tab:failure_llm}
\vspace{3mm}
\end{table}

As discussed in Section~\ref{dataset}, we passed 205 events ($205 \times 4=820$ narratives from Left, Right, Center, and Theme) to LLMs to extract facets. We observed some failures of LLMs while retrieving facets and the failure rate was quite high for Bard (Table~\ref{tab:failure_llm}). For some narratives, the output for Bard was, "\textit{Sorry, as an AI language model, I don't have any knowledge about 5W1H facets}", "\textit{Sorry, I can't help you with that}", "\textit{Sorry, I am new to Language and still learning}" and sometimes Bard threw a null error even after making 3-4 attempts. For ChatGPT, we got empty output occasionally or a message such as "\textit{Sorry, I can't help you with that}".

\subsection{Meta Evaluation}

\subsubsection{Entity-Specific vs. Descriptive Facets}\label{entity_descrip}

\begin{figure}[!htb]
    \centering    
    \includegraphics[width=\linewidth,trim={10 0 10 10},clip]{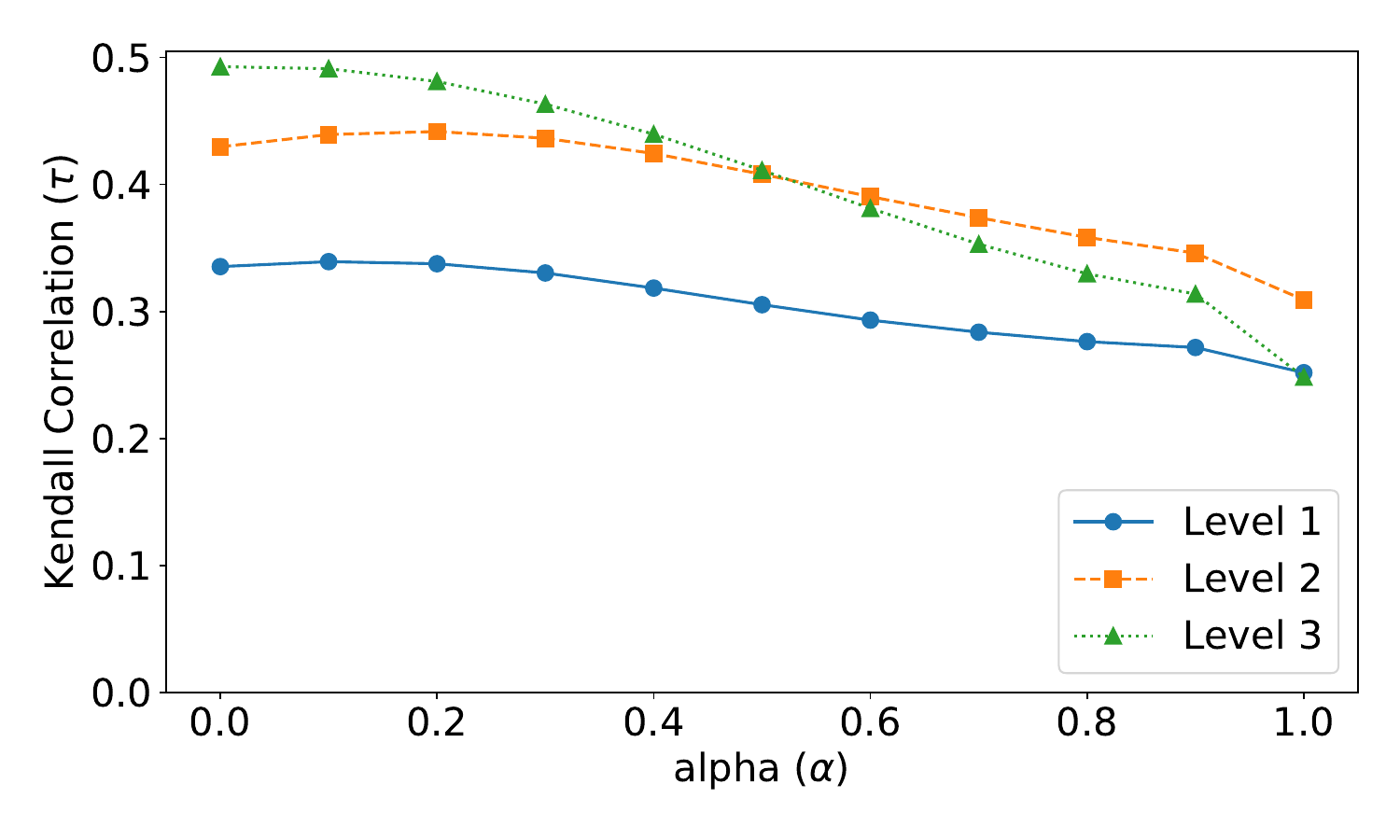}
    
    \caption{Kendall ($\tau$) correlation coefficient when alpha ($\alpha)$ $\in [0, 1]$ for Level 3, Level 2, and Level 1 prompting to retrieve 5W1H facets using ChatGPT. Here, alpha ($\alpha)$ denotes weight for entity-specific facets (\textit{who, when, where}), and (1- $\alpha$) weight is given to descriptive facets (\textit{what, why, how}).}
    \label{fig:level-corr}
    \vspace{-2mm}
\end{figure}

In our experimental setup, we divided the 5W1H facets into two groups: entity-specific facets (who, when, where) and descriptive facets (what, why, how). To compute a final score, we combined these six facets linearly, assigning an alpha ($\alpha$) weight to the entity-specific facets and ($1 - \alpha$) weight to the descriptive facets. This approach allowed us to examine the impact of different alpha values on the correlation against ground truth labels, as demonstrated in Figure~\ref{fig:level-corr}. Here,  we present the Kendall ($\tau$)~\cite{kendall1938new} correlation between the final FaNS score and ground truth labels for variations of $\alpha$ on Level-1/2/3 prompts. 

%Notably, we considered instance-level correlation for our analysis. Conversely, narrative similarity using level 1 prompts exhibited a lower correlation.

Figure~\ref{fig:level-corr} indicates that, for ChatGPT, Level 3 prompts yielded a higher correlation than Level 1 or 2 prompts, which aligns with expectations as Level 3 prompts provide detailed, step-by-step instructions. Interestingly, for $\alpha \geq 0.5$, narrative similarity using Level-2 prompts showed a better correlation than Level-3 prompts. The figure suggests that descriptive facets place greater importance on correlation. To gain further insights into the impact of each facet, we also evaluated the correlation for individual facets, which is discussed in Section~\ref{facet_weight}.

\subsubsection{Examining Individual Facet Impact}\label{facet_weight}
To assess the correlation of individual facets for different levels of prompts, we conducted a facet-wise Kendall ($\tau$) correlation analysis (see Table~\ref{tab:key_corr}). This analysis allowed us to determine the importance of each facet in relation to the narrative.

\begin{table}[!htb]
\centering
\resizebox{\linewidth}{!}{
\begin{tabular}{l|l|c|c|c|c} \hline
 & \textbf{Facet} & \textbf{Level 3} & \textbf{Level 2} & \textbf{Level 1} & \textbf{Average} \\ \hline
& Who & 0.242 & 0.314 & 0.262 & 0.273 \\
\textbf{Entity-}  & When & 0.123 & 0.201 & 0.149 & 0.158 \\
\textbf{Specific} & Where & 0.151 & 0.161 & 0.124 & 0.145 \\ \hline
\multirow{3}{*}{\textbf{Descriptive}} & What & 0.470 & 0.403 & 0.370 & \textbf{0.414} \\
 & Why & 0.378 & 0.291 & 0.145 & 0.271 \\
 & How & 0.293 & 0.268 & 0.184 & 0.248\\\hline
\end{tabular}
}
\caption{Kendall ($\tau$) correlation coefficients on individual facet level similarity for different levels of prompting on ChatGPT. Data-driven order of facets are What $>$ Who $>$ Why $>$ How $>$ When $>$ Where.}
\label{tab:key_corr}
\vspace{3mm}
\end{table}

Table~\ref{tab:key_corr} shows that the \textit{what} facet exhibited a correlation of 41\% with the ground truth labels, indicating its significant role in determining narrative similarity. The \textit{who} facets followed closely as the second highest correlated facet. It is important to note that our analysis was based on the AllSides dataset, which groups narratives according to exact event details. Hence, this finding is aligned with the content-driven nature of the dataset.

Further examination of Kendall's ($\tau$) correlation coefficients on individual facets indicated the following order of importance: What $>$ Who $>$ Why $>$ How $>$ When $>$ Where. These findings provide valuable insights into the relative significance of different facets in measuring narrative similarity.

\subsubsection{Comparison with Existing Metrics}

Our FaNS metric extracts and compares 5W1H facets as a robust way of quantifying narrative similarity, which hasn't been used in any prior work. Therefore, as baselines, we directly computed the similarity of the two narratives using the popular ROUGE~\cite{lin-2004-rouge} and BERTScore~\cite{DBLP:conf/iclr/ZhangKWWA20} metrics and then computed their Kendall ($\tau$) correlation against the ground truth labels.

For the FaNS metric, we present correlation results for two different approaches. In the first approach, we assigned a weight of 0.2 to entity-specific facets and 0.8 to descriptive facets (based on our findings from Figure~\ref{fig:level-corr}). In the second approach, we employed a different weighting scheme where the weights were directly derived from the facet-wise correlation numbers as discussed in Section~\ref{facet_weight}. More specifically, the weight of a particular facet was set in proportion to its individual correlation with the ground truth, i.e., facets with higher correlation were assigned higher weights.

Results from Table~\ref{tab:key_comb_corr} demonstrate that both approaches exhibited significantly higher correlation compared to BERTScore for both Level 3 and Level 2 prompts. Specifically, the linear weighting of entity-specific and descriptive facets at Level 3 demonstrated approximately \textbf{37\%} higher correlation than BERTScore. Similarly, for the correlation-driven-weighted average approach with Level 2 prompts, we observed a \textbf{15\% }increase in correlation compared to BERTScore. In contrast, ROUGE showed poor correlation across all samples, reaffirming its limitations in capturing similarity for complex narratives due to its sole reliance on the lexical overlap.

\begin{table}[!htb]\small
\centering
\resizebox{0.4\textwidth}{!}{
\begin{tabular}{l|c|c|c}
\hline
\textbf{Metric} & \textbf{Level 3} & \textbf{Level 2} & \textbf{Level 1} \\ \hline
ROUGE-1 & -0.02 & 0.02 & -0.03 \\ 
ROUGE-2 & 0.19 & 0.2 & 0.19 \\ 
ROUGE-L & 0.03 & 0.07 & 0.03 \\ 
BERTScore & 0.35 & 0.39 & \textbf{0.36} \\ \hline
FaNS\textsuperscript{\textbf{*}} & \textbf{0.48} & 0.44 & 0.34 \\ 
FaNS\textsuperscript{\textbf{†}} & 0.45 & \textbf{0.45} & 0.35 \\ \hline
\end{tabular}
}
\caption{Kendall ($\tau$) correlation coefficients where FaNS\textsuperscript{\textbf{*}} denotes $\alpha=0.2$ for entity-specific facets. For FaNS\textsuperscript{\textbf{†}}, weights of facets are set in proportion to their correlation scores (What $>$ Who $>$ Why $>$ How $>$ When $>$ Where). ChatGPT is used for extracting facets for both versions of FaNS. The highest correlation value is bolded in each column.}
\label{tab:key_comb_corr}

\end{table}

\section{Conclusion}
In this paper, we have introduced a facet-based narrative similarity (FaNS) metric based on the classic 5W1H facet representation of events. By leveraging the power of Large Language Models (LLMs) like ChatGPT and Bard, we first structured narratives in terms of these facets, which was a challenging task before the LLM era. Next, we computed the similarity of each facet individually and combined them to provide a final similarity score. By employing facet-based similarity metrics, FaNS provides better control over the evaluation process and enhances the explainability of the results.

To meta-evaluate our metric, we have curated a comprehensive dataset by collecting narratives from AllSides and heuristically annotated this data set with ground truth labels. Our experiments demonstrate that the facet-based narrative similarity metric exhibits a higher correlation (\textbf{37\% $\uparrow$}) against the ground-truth labels than directly measuring the similarity between narratives through traditional metrics like ROUGE and BERTScore. This indicates that considering the 5W1H facets of a narrative leads to a more accurate assessment of narrative similarity.

We believe that our work provides a valuable foundation for future research in the field of narrative understanding and similarity measurement by contributing a new data set and an evaluation metric. We encourage researchers to use the FaNS metric for measuring narrative similarity scores in future works.

%Overall, our contributions include the proposal of comparing narratives along the 5W1H facets for the development of the facet-based narrative similarity (FaNS) metric and the creation of a dedicated dataset for evaluating narrative similarity. 

\section{Limitation}
There are some limitations of our work that should be considered. Firstly, we conducted our experiments solely within the news domain, and there might be various challenges associated with applying facet-based similarity metrics to other domains. Thus, further investigation is needed to explore the applicability and effectiveness of our approach in other domains. Additionally, as a baseline for comparison, we only considered the ROUGE and BERTScore metrics. It is worth noting that there are numerous metrics available for evaluating text summarization/generation, but our study does not focus on exploring these metrics extensively rather we focus on narrative similarity. Furthermore, it is worth highlighting that there are currently no existing metrics that can directly serve as a baseline for our work, as previous research has not developed any metric specifically based on the 5W1H facets.

Another important limitation of the proposed FaNS metric is that its accuracy greatly depends on the quality of facets extracted by the LLMs. However, our results suggest that LLMs, like ChatGPT, are quite robust and can indeed extract facets with reasonable accuracy.

\section{Ethics Statement}
We utilized publicly accessible news articles as the primary source for constructing the dataset for the experiments. As a result, to the best of our knowledge, there are no ethical violations. Additionally, the evaluation of narrative similarity is a major aspect of this work. Hence, we consider it a low-risk research study.

% \clearpage
\bibliography{references}
\bibliographystyle{acl_natbib}
\appendix
\section{Appendix}
\subsection{Experiment with Google Bard}
\subsubsection{Entity-Specific vs. Descriptive Facets: Correlation Impact}
\begin{table}[!htb]
\centering
\resizebox{0.48\textwidth}{!}{
\begin{tabular}{cc|cc|cc}
\hline
\multicolumn{2}{c|}{\textbf{Level 1}} & \multicolumn{2}{c|}{\textbf{Level 2}} & \multicolumn{2}{c}{\textbf{Level 3}} \\ \hline
\multicolumn{1}{l|}{\textbf{Score}} & \multicolumn{1}{l|}{\textbf{Pairs (\#)}} & \multicolumn{1}{l|}{\textbf{Score}} & \multicolumn{1}{l|}{\textbf{Pairs (\#)}} & \multicolumn{1}{l|}{\textbf{Score}} & \multicolumn{1}{l}{\textbf{Pairs (\#)}} \\ \hline
\multicolumn{1}{c|}{3} & 472 & \multicolumn{1}{c|}{3} & 498 & \multicolumn{1}{c|}{3} & 454 \\ 
\multicolumn{1}{c|}{2} & 446 & \multicolumn{1}{c|}{2} & 491 & \multicolumn{1}{c|}{2} & 441 \\ 
\multicolumn{1}{c|}{1} & 500 & \multicolumn{1}{c|}{1} & 500 & \multicolumn{1}{c|}{1} & 500 \\ 
\multicolumn{1}{c|}{0} & 500 & \multicolumn{1}{c|}{0} & 500 & \multicolumn{1}{c|}{0} & 500 \\ \hline
\end{tabular}
}
\caption{Statistics of the dataset used to calculate correlation in Google Bard}
\label{tab:data_stat_bard}
\end{table}

\begin{figure}[!htb]
    \centering
    \begin{subfigure}[b]{0.48\textwidth}
        \centering
        \includegraphics[width=\linewidth,trim={10 0 10 10},clip]{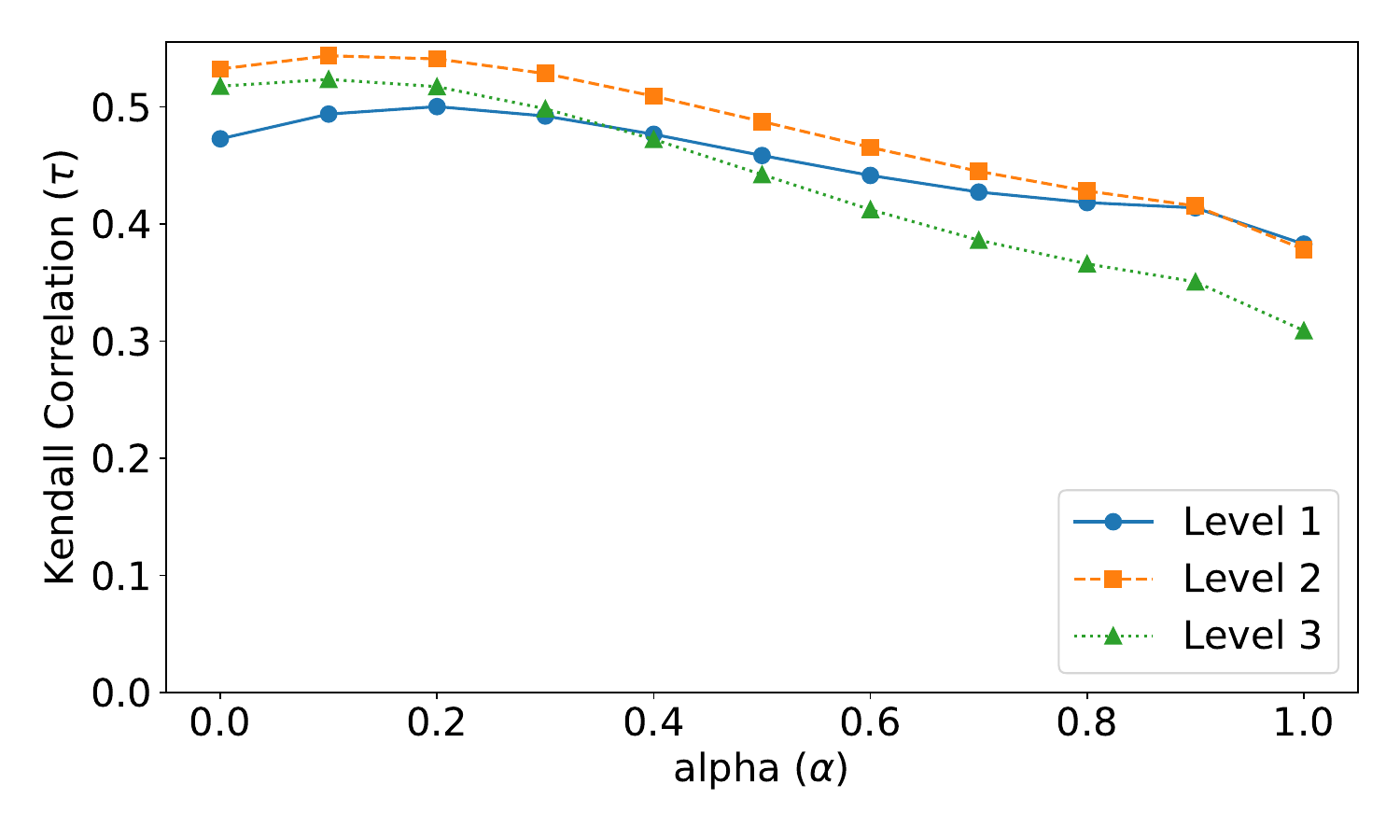}
    \end{subfigure}
    \caption{Kendall ($\tau$) Correlation coefficient when alpha ($\alpha)$ $\in [0, 1]$ for Level 3, Level 2, and Level 1 prompting to retrieve 5W1H facets using Google Bard. Here, alpha ($\alpha)$ denotes weight for entity specifics (\textit{who, when, where}) and (1- $\alpha$) weight is given to descriptive specifics (\textit{what, why, how}).}
    \label{fig:level-corr-bard}
\end{figure}

Google Bard~\cite{bard} experienced a high failure rate, resulting in a reduced number of samples available for correlation calculations compared to ChatGPT (see Table~\ref{tab:data_stat_bard}). Figure~\ref{fig:level-corr-bard} shows that the highest correlation for Google Bard is achieved when utilizing a linear combination of entity-specific and descriptive weights, particularly when the $\alpha$ value is set to 0.2 for Level 2 prompting. This indicates that allocating 20\% weight to entity-specific facets and 80\% weight to descriptive facets yields the strongest correlation while using Google Bard for facet retrieval. Interestingly, for an $\alpha$ value of 0.4 Level 1 prompting leads to a better correlation than Level 3 prompting. Furthermore, when $\alpha$ is set to 0.9, level 1 and level 2 prompting exhibit similar correlations. These findings suggest that Google Bard struggles to capture nuanced differences in instruction.

\subsubsection{Examining Facet Importance in Narrative: Which Facet Drive Correlation?}
From Table~\ref{tab:key_corr_bard} we can see that \textit{What} facet still exhibits the highest correlation when utilizing Google Bard. Interestingly, the data-driven order of facets in terms of correlation strength with Google Bard is  What $>$ Why $>$ Who $>$ How $>$ Where $>$ When. This order is somewhat similar to the correlation pattern observed with ChatGPT.

\setlength{\textfloatsep}{0pt}

\begin{table*}[!htb]
\centering
\resizebox{0.55\textwidth}{!}{
\begin{tabular}{l|l|c|c|c|c} \hline
 & \textbf{Facets} & \textbf{Level 3} & \textbf{Level 2} & \textbf{Level 1} & \textbf{Average} \\ \hline
\multirow{3}{*}{\textbf{Entity-Specific}} & Who & 0.298 & 0.353 & 0.398 & 0.350 \\
 & When & 0.205 & 0.247 & 0.248 & 0.233 \\
 & Where & 0.248 & 0.297 & 0.29 & 0.278 \\ \hline
\multirow{3}{*}{\textbf{Descriptive}} & What & 0.481 & 0.487 & 0.406 & \textbf{0.458} \\
 & Why & 0.376 & 0.389 & 0.315 & 0.360 \\
 & How & 0.306 & 0.405 & 0.282 & 0.331\\\hline
\end{tabular}
}
\caption{Kendall ($\tau$) correlation coefficients on individual facet level similarity for different levels of prompting on Google Bard. Data-driven order of facets are What $>$ Why $>$ Who $>$ How $>$ Where $>$ When.}
\label{tab:key_corr_bard}
\end{table*}

\subsubsection{Comparison with Baseline Metric}
When comparing the performance of FaNS with baseline ROUGE and BERTScore metrics, it becomes evident that calculating narrative similarity with BERTScore using two narratives yields a high correlation. This outcome emphasizes the failure of Google Bard in retrieving 5W1H facets, which is a crucial aspect of our FaNS  metric. It is important to note that the reported number cannot be directly compared with Table~\ref{tab:key_comb_corr} due to differences in sample sizes.

\begin{table*}[!htb]\small
\centering
\resizebox{0.4\textwidth}{!}{
\begin{tabular}{l|c|c|c}
\hline
\textbf{Metric} & \textbf{Level 3} & \textbf{Level 2} & \textbf{Level 1} \\ \hline
ROUGE-1 & 0.32 & 0.33 & 0.35 \\
ROUGE-2 & 0.45 & 0.46 & 0.46 \\
ROUGE-L & 0.34 & 0.34 & 0.36 \\
BERTScore & \textbf{0.56} & \textbf{0.54} & \textbf{0.58} \\\hline
FaNS\textsuperscript{\textbf{*}} & 0.52 & \textbf{0.54} & 0.5 \\
FaNS\textsuperscript{\textbf{†}} & 0.50 & 0.53 & 0.50\\ \hline
\end{tabular}
}
\caption{Kendall ($\tau$) correlation coefficients where FaNS\textsuperscript{\textbf{*}} denotes $\alpha=0.2$ for entity-specific weights and $1-\alpha=0.8$ weight for descriptive facets. On the other hand, weights of facets are ordered based on data-driven priority (What $>$ Why $>$ Who $>$ How $>$ Where $>$ When) in FaNS\textsuperscript{\textbf{†}} for different levels of prompting on Google Bard. The highest correlation value is bolded in each column.}
\label{tab:key_comb_corr_bard}
\end{table*}

\subsection{Explanation of Metrics/Packages} \label{metric}
\noindent{\bf {ROUGE~\cite{lin-2004-rouge}: }} Between a pair of narratives, ROUGE\footnote{\href{https://huggingface.co/spaces/evaluate-metric/rouge}{https://huggingface.co/spaces/evaluate-metric/rouge}} counts the overlap of textual units (n-grams, word sequences).

\noindent{\bf {BERTScore~\cite{DBLP:conf/iclr/ZhangKWWA20}:}} BERTScore\footnote{\href{https://huggingface.co/spaces/evaluate-metric/bertscore}{https://huggingface.co/spaces/evaluate-metric/bertscore}} calculates embedding based similarity scores by matching narrative pairs on a token level. The cosine similarity between contextualized token embeddings from BERT is maximized by computing token matching greedily.

\noindent{\bf {ConceptNet~\cite{DBLP:conf/aaai/SpeerCH17}:}} ConceptNet\footnote{\href{https://conceptnet.io/}{https://conceptnet.io/}} is a knowledge graph that represents general knowledge using nodes (concepts) and edges (relationships), making inferences based on semantic connections between concepts.

\noindent{\bf {Python dateutil\footnote{\href{https://dateutil.readthedocs.io/en/stable/parser.html}{https://dateutil.readthedocs.io/en/stable/parser.html}}:}} is a Python library that provides a powerful date and time parsing, manipulation, and arithmetic functionalities. It allows easy parsing of dates from strings in various formats, handling ambiguous dates, time zone conversions, and calculating time differences.

\noindent{\bf {Spacy NER\footnote{\href{https://spacy.io/usage/embeddings-transformers}{https://spacy.io/usage/embeddings-transformers}}:}} performs entity extraction by identifying and classifying named entities in text. It also supports embedding extraction enabling downstream NLP tasks.

\subsection{Computational Infrastructure}
For experiments, we used NVIDIA Quadro RTX 5000 GPUs.

\end{document}